\documentclass{article}

\usepackage{arxiv}

\usepackage[utf8]{inputenc} % allow utf-8 input
\usepackage[T1]{fontenc}    % use 8-bit T1 fonts
\usepackage{hyperref}       % hyperlinks
\usepackage{url}            % simple URL typesetting
\usepackage{booktabs}       % professional-quality tables
\usepackage{amsfonts}       % blackboard math symbols
\usepackage{nicefrac}       % compact symbols for 1/2, etc.
\usepackage{microtype}      % microtypography
\usepackage{graphicx}
\usepackage{mathtools}
\usepackage{amsmath}
\usepackage{algorithm}
\usepackage{algpseudocode}
\usepackage{subfig}
\usepackage{multirow}
\newcommand{\defeq}{\coloneqq}
\DeclarePairedDelimiter\ceil{\lceil}{\rceil}

\DeclareMathOperator*{\argmax}{arg\,max}

\title{Logographic Subword Model for Neural Machine Translation}

\author{
  Yihao Fang \\
  Department of Computing and Software \\
  McMaster University \\
  Hamilton, ON L8S 4L8 \\
  \texttt{fangy5@mcmaster.ca} \\
   \And
  Rong Zheng \\
  Department of Computing and Software \\
  McMaster University \\
  Hamilton, ON L8S 4L8 \\
  \texttt{rzheng@mcmaster.ca} \\
   \And
  Xiaodan Zhu \\
  Department of Electrical and Computer Engineering \\
  Queen's University \\
  Kingston, ON K7L 3N6 \\
  \texttt{xiaodan.zhu@queensu.ca} \\
}

\begin{document}
\maketitle

\begin{abstract}
A novel logographic subword model is proposed to reinterpret logograms as abstract subwords for neural machine translation. Our approach drastically reduces the size of an artificial neural network, while maintaining comparable BLEU scores as those attained with the baseline RNN and CNN seq2seq models. The smaller model size also leads to shorter training and inference time. Experiments demonstrate that in the tasks of English-Chinese/Chinese-English translation, the reduction of those aspects can be from $11\%$ to as high as $77\%$. Compared to previous subword models, abstract subwords can be applied to various logographic languages. Considering most of the logographic languages are ancient and very low resource languages, these advantages are very desirable for archaeological computational linguistic applications such as a resource-limited offline hand-held Demotic-English translator. 
\end{abstract}


\section{Introduction}
Sequence-to-sequence (seq2seq) models are widely used in neural machine translation. Recurrent neural network (RNN) based sequence-to-sequence models were first proposed by Cho et al. \cite{Cho:14} and Sutskever et al. \cite{Sutskever:14}. The models have gained extensive attention and motivated many efforts to make them faster, smaller, and more accurate, such as tied embedding \cite{Press:16}, layer normalisation \cite{Ba:16}, weight normalisation \cite{Salimans:16}, and subword byte pair encoding (BPE) \cite{Sennrich:16b,Gage:94}, among others. The convolutional neural network (CNN) based sequence-to-sequence models \cite{Gehring:17} substitute RNN components with CNN and allow much faster training while retaining the BLEU scores closely comparable to those obtained with RNN seq2seq models.

Despite the success of seq2seq models in machine translation, their high computing complexity still strongly limits their applications such as those running on offline hand-held translators. An efficient approach to reducing models' complexity is compacting their output layers. A cumbersome output layer significantly increases the number of parameters in the model and consequently slows down model inference. Furthermore, the amount of gradient calculation in training grows as the number of model parameters increases. In machine translation, the size of the output layer is often proportional to the size of the target dictionary. For instance, if there are one million words in the target dictionary, $64$ examples in a batch, and $50$ words in a sentence, then there are $64 \times 1M$ units in the output layer of each RNN decoder cell, and $64 \times 50 \times 1M$ units in the output layer of a CNN decoder. Compacting the target dictionary consequently reduces the model size and speeds up model training and inference. 

It is non-trivial to design a new approach that can compact the target dictionary and is directly applicable to different logographic languages without sacrificing performance. In this paper, we propose a logographic subword model that represents logograms as multiple ``abstract subwords'' (code symbols), with an encoder and decoder transforming logograms to abstract subwords and subwords to logograms. The encoder quantizes and decomposes the embeddings of logograms to multiple abstract subwords (code symbols). Word embedding or equivalent vector representation of words \cite{Pennington:14,Mikolov:13} preserves the closeness between word pairs through their distances in the vector space. Quantization identifies common semantic components (abstract subwords) among logograms. Two quantizers are examined in the experiments: the state of the art locally optimized product quantization (LOPQ) \cite{Kalantidis:14} and a novel density aware product quantization (DAPQ). Quantizing embeddings of logograms helps logograms with close meanings be more likely to share common abstract subwords in one or more dimensions. By sharing, the number of abstract subwords can be significantly reduced in the target directory. Furthermore, only infrequent words are decomposed into code symbols. Taking word frequency into consideration avoids unnecessary elongation of the source and target sentences.

Using the proposed model, the sizes of RNN and CNN sequence-to-sequence models are reduced by $37\%$ and $77\%$ respectively in an English-to-Chinese translation task without sacrificing performance. The training times are about $11\%$ and $73\%$ shorter; the inference time is nearly halved in RNN and $36\%$ shorter in CNN.

Considering many of the logographic languages are ancient and low resource languages, these advantages are also desirable for archaeological applications. Also, reduction in model sizes is useful for resource-limited applications such as those running on hand-held devices. Furthermore, while we discuss the proposed models in the context of translation, the methods have implications for other tasks of NLP involving predicting tokens, e.g., language modeling, summarization, and image captioning.

The rest of the paper is organized as follows. Section 2 describes the related works in neural machine translation. The proposed approach is presented in Section 3. Section 4 provides more details about product quantization and the proposed density aware approach. In Section 5, we discuss more the encoding algorithm. The experimental setups and experimental results are given in Section 6, and conclusions and future work are discussed in Section 7.

\section{Related Works}
\label{sec:related_works}
Cho et al \cite{Cho:14} first proposed the RNN seq2seq model by modeling it as an RNN encoder-decoder architecture, with the encoder transforming an input sentence into a context vector and the decoder mapping the context vector to an output sentence (the translation hypotheses). Bahdanau et al. \cite{Bahdanau:14} further improved RNN seq2seq models by making the encoder as a bidirectional gate recurrent unit (GRU) and binding the attentional mechanism to the decoder. To avoid overfitting in RNN seq2seq models, Gal and Ghahramani \cite{Gal:16} proposed to apply the variational inference based dropout technique to the model. To speed up convergence in training, Ba and his colleagues \cite{Ba:16} introduced layer normalization to stabilize state dynamics in RNNs. Salimans and Kingma \cite{Salimans:16} proposed weight normalization that reparameterizes weight vectors from their direction. To increase the model depth, Zhou et al. \cite{Zhou:16} proposed fast-forward connections where the shortest paths do not depend on any recurrent calculations. Wu et al. \cite{Wu:16} introduced the bidirectional stacked encoder and Barone and his colleague \cite{Barone:17} proposed a BiDeep RNN by replacing the GRU cells of a stacked encoder with multi-layer transition cells. Deviating from RNN seq2seq models, Gehring et al. \cite{Gehring:17} proposed convolutional seq2seq model where the encoder and decoder were fully replaced by convolutional neural networks (CNN). Their approach allows much faster training while retaining the BLEU scores closely comparable to those obtained with RNN seq2seq models. 

\paragraph{Embedding}
Sennrich and Haddow \cite{Sennrich:16a} generalized the embedding layer to support linguistic features such as morphological features, part-of-speech tags, and syntactic dependency labels. Press and Wolf \cite{Press:16} proposed tier embedding and argued that weight tying reduces the size of neural translation models. However, no attempt has been made to reduce the size of the target dictionary through word embedding. Significant reduction in model complexity is expected considering words are on the order of hundreds of thousands or more in a typical dictionary. 

\paragraph{Decomposition}
Sennrich et al. \cite{Sennrich:16b} proposed to segment words of source and target sentences into smaller subword units using byte pair encoding (BPE) compression \cite{Gage:94}. They showed an improvement in the BLEU scores of 1.1 and 1.3 for English-German and English-Russian translations, respectively. Despite its advantage, BPE splits an alphabetic word to multiple letter groups, and thus it is intrinsically not applicable to logographic languages such as Chinese, Chorti, and Demotic (Ancient Egyptian) where a word is a glyph rather than alphabetic letters. Garc{\'\i}a-Mart{\'\i}nez et al. \cite{Garcia-Martinez:16} proposed to decompose words morphologically and grammatically into factored representations such as lemmas, part-of-speech tag, tense, person, gender, and number. Their approach reduced training time and out of vocabulary (OOV) rates with improved translation performance, but also introduces unnecessary grammatical dependencies, (e.g. there are hundreds of tenseless languages), and is not optimized in all scenarios. 

To the best of our knowledge, our work is the first to explore an abstract subword representation for logographic languages. The abstract representation makes it applicable to different logographic languages. It is a very desirable feature especially for most of the low resource logographic languages.

\section{System Architecture}

\begin{figure}[t]
\centering
\includegraphics[width=0.5\linewidth]{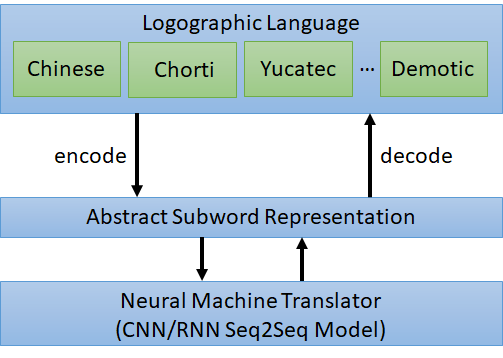}
\caption[Architecture Diagram]{Abstract subwords are code symbols which are independent on a particular language. They directly participate in machine translation on behave of words of the logographic language. }
\label{fig:approach}

\end{figure}

The proposed logographic subword model consists of an encoder and a decoder. The encoder transforms a word into multiple abstract subwords (code symbols), and the decoder transforms multiple abstract subwords into a word (Figure \ref{fig:approach}). Only abstract subwords directly participate in the training of the sequence-to-sequence models. This additional layer of abstraction reduces the model size, because the smaller dictionary that the abstract subwords form results in an output layer with the smaller number of units in the neural network. 

\begin{figure}[t]
\centering
\includegraphics[width=0.45\linewidth]{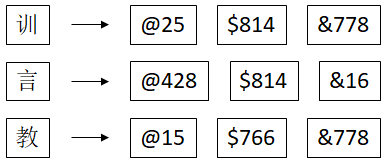}
\caption{The encoder converts the word ``train'' in the corpus to three new abstract subwords (symbols) ``@25'', ``\$814'' and ``\&778''. ``Train'' and ``talk'' share the second symbol ``\$814'', and ``train'' and ``teach'' share the third symbol ``\&778''. Through sharing, the number of distinct symbols in the dictionary can be greatly reduced. }
\label{fig:examples} 
\end{figure}

Abstract subwords are code symbols which are independent on a particular language. These symbols are shared among words, and thus there will be much fewer symbols in both the source and target dictionaries (Figure \ref{fig:examples}). Symbols are created from a quantizer's codebook. The code at the first dimension is left padded an ``@'' and the code at the second dimension is left padded a ``\$'', etc.

%The decoder converts code symbols back to words. Since only code symbols are involved in training, the classes predicted by the model are code symbols. Code symbols are then replaced by their coordinates in word embedding. The nearest words to the coordinate vectors then form the final translated sentences. Finding the nearest word has the benefit of mitigating potential errors introduced by neural networks. 

\subsection{Encoder}

The encoder quantizes and decomposes the embeddings of words into abstract subwords. Embedding maps words to vectors of real numbers. Similarities among words in the corpus are reflected to a large extent by Euclidean distances, which preserve the relationship between pairs of words. However, machine translation is generally modeled as a classification problem rather than a regression problem. Consequently, vectors of real numbers need to be represented by vectors of distinct symbols. This can be accomplished via quantization. 

Quantization transforms vectors of real numbers to code symbols. In contrast to the initial word identifiers (before embedding), code symbols embed the information about word similarities. Furthermore, the code symbols can be far fewer than the word identifiers, and this alleviates the classification burden and reduces the size of the neural network. To retain the translation performance, quantization also needs to enforce a one-to-one correspondence between words and vectors of code symbols.

In order not to significantly increase the length of a long sentence, the decomposition procedure ensures that only infrequent words are replaced by their quantized vectors of code symbols. Code symbols (abstract subwords) participate in the training of the sequence-to-sequence model, as opposed to word identifiers. 

The encoder transforms words in the corpus to fewer distinct symbols (abstract subwords) resulting in a smaller dictionary.  A smaller target dictionary leads to a smaller model size. With the smaller number of weights and bias, less gradient calculation is needed, and training is sped up. On the other hand, with the smaller number of weights and bias, there are fewer operations during inference, and inference time is shortened. 

More details of quantization and decomposition will be covered in section 4 and section 5 respectively.

\subsection{Decoder}
Since code symbols are used in training, actual words are unknown to the neural network. Inference outputs are sequences of code symbols (abstract subwords). Symbol-level prediction errors may make it impossible to exactly locate a word by the sequence of decoded symbols. We notice that most of the time, such errors tend to be minor, with only one of symbols in the sequence incorrectly predicted. In order to restore the original symbols and locate the right word, we can apply the nearest neighbor search \cite{Jegou:11} to efficiently decode code symbols to words of the logographic language even in presence of prediction errors. 

Experiments show that the translation performance (BLEU score) with our approach actually matches that without any preprocessing, since the nearest neighbor search to some extent rectifies prediction errors, and with the smaller number of output classes (neurons), the resulting neural network has lower complexity.

\section{Product Quantization}

Quantization is an important component of the encoder. In our experiments, product quantization methods are examined. Product quantization was first proposed by J\'egou and his colleagues \cite{Jegou:11} for nearest neighbor search. They decomposed the space into lower dimension subspaces and perform quantization on each subspace separately. Consider a space of $n$ dimensions evenly divided into $m$ subspaces each of $n/m$ dimensions. Quantization is then performed on each of the $m$ subspaces separately and maps vectors in each subspace to codes in each sub-codebook.

Formally, let $ X $ be the set of all vectors in the space, and $x$ is a vector in $X$. Assume that $x$ can be evenly divided into $m$ subvectors, noted by $ u_i(x)$, where $i \in I = \{1,...,m\}$. We have that $x$ is the concatenation of all $u_i(x)$, noted by 
\begin{displaymath}
x = u_1 (x)||...||u_m (x) ,
\end{displaymath}
where symbol $ || $ stands for concatenation. With the above definitions, subspace $X_i$ can be defined as the set of all $u_i(x)$ for all $x \in X$, noted by
\begin{displaymath}
X_i \defeq \{u_i(x)|x \in X\}, \forall i \in I
\end{displaymath}
The $m$ subspaces are quantized separately. Subquantizer $q_i$ of subspace $X_i$ maps $u_i(x)$ to a reference vector in $C_i = \{c_{i,1}, c_{i,2}, \ldots, c_{i,k_i}\}$, where $k_i$ is the size of the codebook $C_i$ for $X_i$. Here, $q_i$ defines a partition of $X_i$, namely, 
\begin{displaymath}
S_{i,j} \defeq \{u_i(x)|x \in X  \text{ and } q_i(u_i(x)) = c_{i,j}\}, 
\end{displaymath}
where $i \in I, j \in J_i = \{1, 2, \ldots, k_i\}$. 

Let $q(x)$ be the concatenation of all the reference vectors, noted by
\begin{displaymath}
q(x) = q_1(u_1(x)) ||...||q_m(u_m(x))
\end{displaymath}
The quality of a quantizer is usually measured by the mean squared error between the input vector $x$ and its reproduction value $q(x)$, 
\begin{displaymath}
E_{PQ} = \sum_{x \in X} || x - q(x)||^2
\end{displaymath}
For a quantizer to be optimal, it should quantize the subvectors of $x \in X$ to their nearest centroids, namely, 
\begin{displaymath}
c_{i,j}=\frac{1}{|S_{i,j} |} \sum_{u_i(x) \in S_{i,j}} u_i(x)
\end{displaymath}

\subsection{Centroid Initialization and Gaussian Kernel Density} \label{subsec:quantization}

Product quantization \cite{Jegou:11} is an effective quantization method. However, the design of the quantizer is application dependent. With respect to the GloVe \cite{Pennington:14} or Word2Vec \cite{Mikolov:13} vector spaces, our experiments show that higher frequency words tend to have higher densities around in the space. Since high-frequency words are more likely to appear in the target sentences, intuitively, making higher frequency words more distinguishable helps improving translation performance. Thus, we propose a new density aware product quantization (DAPQ), which is a variant of product quantization customized for word embedding and machine translation.  

First, we estimate the Gaussian kernel density for every subspace. The log density is used in calculating the initial centroids for the k-means++ algorithm \cite{Arthur:07}. Initially, the first centroid is selected uniformly at random from $X_i$, and subsequent centroids are chosen from any vector $z$ in $ X_i$ with probability: 
\begin{equation} \label{eq:rho}
p_i(z)=\frac{\rho(z) D (z)^2}{\sum_{z \in X_i} \rho(z) D(z)^2}  , z \in X_i , i \in I ,
\end{equation}
where $\rho(z)$ is the Gaussian kernel density of $z$ and $D(z) $ denotes the distance from $z$ to its nearest centroid. The density term $\rho(z)$ allows more centroids to be initially assigned to denser areas. Consequently, dense words, which are also frequent words, are more likely to be grouped into distinct clusters in subspaces. Doing so increases the chances frequent words being correctly decoded in the target sentences.

\subsection{The Number of Clusters and Degree of Distinctness}

Quantization locally minimizes the distortion in each subspace. However, there are a number of global hyperparameters that remain to be decided. In product quantization, there are two important hyperparameters: \textit{the number of subspaces} $m$ and \textit{the number of clusters in each subspace} $k$. It is observed that the selections of $m$ and $k$ affect the translation performance (BLEU scores), but their effects are indirect and not analytically tractable. However, we observe that a more distinct correspondence (e.g. the most distinct one-to-one correspondence) tends to improve translation performance. Thus, it is necessary to introduce a metric to measure the \textit{degree of distinctness} DoD (Figure \ref{fig:CNN-DoD-BLEU}). Intuitively, larger $k$ tends to increase the distinctness and consequently translation performance. Therefore, it is reasonable to use DoD as a performance metric to find the best number of clusters $k$ in each subspace.

\subsubsection{Degree of Distinctness}

DoD affects the reversibility of quantization. It actually tells how unique the code vectors are. Uniqueness is critical, e.g., if both ``mile'' and ``kilometer'' are encoded to the same code vector, ambiguity will arise. It is not reversible no matter selecting either one (e.g., based on weights), and it will hurt the translation quality. Our experiment demonstrates that the degree of distinctness (uniqueness) actually significantly impacts the final BLEU score in machine translation. Thus, it is a crucial metric to consider in designing the encoding algorithm.

%There might not be a one-to-one correspondence between a word and its code vector. Our experiments demonstrate that translation performance is correlated to the distinctness of the correspondence. A more distinct correspondence leads to a better translation performance. For instance, if ``mile'' and ``kilometer'' are mapped to the same code vector, they are not differentiable from the translation outputs, and it will affect the translation performance. That is the situation to avoid. It is very necessary to define a metric to quantify the effect of distinctness. We call the metric \textit{degree of distinctness} (DoD).  

Formally, let $W$ be the set of all the words from the logographic dictionary and $Q$ is the set of all the distinct code vectors generated by quantization. DoD $\mathfrak{D}$ is defined as $e$ to the power of the multiplication of a scaler $b$ and the result of $1$ subtracted by the cardinality of set $W$ divided by that of set $Q$, noted by: 
\begin{equation} \label{eq:D}
\mathfrak{D} =  e^{b(1-\frac{|W|}{|Q|})},
\end{equation}
where $|\cdot|$ is the cardinality of a set.

\subsubsection{Optimizing the Number of Clusters in each Subspace}
As the number of clusters increases, DoD increases at the cost of a slightly larger dictionary. Denote $D_{trgt}\in [0,1]$ be the target DoD. To find the value of $k$, starting from an initial value $\ceil{\sqrt[m]{|X|}}$, we incrementally search with step size $\eta$ until the resulting DoD reaches $D_{trgt}$. 

More generally, the number of clusters can be different from subspace to subspace. Let $k_i$ be the number of clusters in subspace $X_i$ with an initial value of $\ceil{\sqrt[m]{|X|}}$. 
Finding the optimal $k_i$'s is an integer programming problem. We adopt a simple heuristic similar to coordinate descent by increasing the number of clusters in one subspace at a time. Specifically, for each $i$, we compute $\Delta D_i$ as the change in DoD when increasing $k_i$ by $\eta$ while keeping the others the same. Let $i^* = \argmax_{i \in I}\Delta D_i$. Then, we update $k_{i^*} = k_{i^*} + \eta$.  The process is repeated until $D_{tgrt}$ is met.

\subsection{Reduction in Target Dictionary}

The target dictionary size is given by $\sum_i k_i$. If $D_{tgrt} = 1$, it is easy to show that $\prod_i k_i \ge |X|$. This implies that $\sum_i k_i \ge m\sqrt[m]{|X|}$. Therefore, the maximum reduction in vocabulary size is $\frac{\sum_i k_i \ge m\sqrt[m]{|X|}}{|X|}$. Empirical results show that $k_i$'s found by the heuristics discussed previously are mostly close to their initial values $\sqrt[m]{|X|}$, resulting significant reduction in dictionary size. For example, for $|X| = 64000$ and $m = 3$, the maximum reduction is $64000/3*40=533$.

\section{Decomposition}
Decomposition is another part of the encoder. Quantizing words to multiple code symbols can significantly reduce the target dictionary size. However, doing so for each word of a sentence would unnecessarily increase the length of the sentence. Experiments show that longer sentences would adversely affect BLEU scores, training and inference times. On the other hand, a smaller dictionary is beneficial to these metrics. Clearly, there exists a trade-off between the dictionary size and sentence length.

To address this problem, the decomposer measures the frequency of words in the corpus, defined as the number of occurrences of a word in the corpus divided by the total number of words in the corpus. It is observed that most words in the corpus are infrequent words, while most sentences are largely composed of frequent words. Therefore, it is sufficient and beneficial to only decompose infrequent words to their vectors of code symbols, (to reduce the dictionary size but not to significant increase sentence length).

\begin{figure}[t]
\centering
\includegraphics[width=0.5\linewidth]{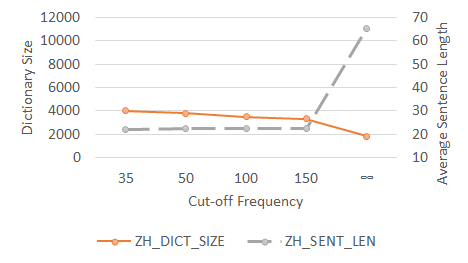}
\caption[Sentence Length]{For the logographic language Chinese, as the cut-off frequency increases, the number of symbols decreases while the average sentence length increases. The BLEU score, training, and inference are benefited from the smaller dictionary and shorter sentences. There is a trade-off between them since they increase/decrease in the opposite directions with respect to the cut-off frequency. }
\label{fig:dataset_stats}

\end{figure}

We introduce parameter $f_{ct}$ as the cut-off frequency to judge whether a word is an infrequent word or not. If a word's frequency is larger than the cut-off frequency, it is a frequent word, or else an infrequent word. As shown in Figure \ref{fig:dataset_stats}, a larger cut-off frequency leads to a smaller target dictionary size, but a longer sentence on average, and vice versa.

In summary, the proposed encoder (as described in Algorithm \ref{alg:encode}) takes as input the set of all words $W$, the corpus of all sentence pairs $C$, the number of partitions $m$, the distinctness objective $\mathfrak{D}_{trgt}$, the learning rate $\eta$, and the cutoff frequency $f_{ct}$. It calculates the (GloVe \cite{Pennington:14}) embedding of $W$ and assigns it to $X$. It initiates the degree of distinctness $\mathfrak{D}$ and the number of clusters in each subspace $k$. Here, $k$ is initialized as $\lceil \sqrt[m]{|X|} \rceil $, and increases iteratively until the target degree of distinctness is reached. Product quantization takes $k$ as the hyper-parameter and outputs the set of code symbols with minimum distortion. Decomposition decomposes only infrequent words to their code symbols without over elongation of the sentences based on the cut-off frequency $f_{ct}$.

\begin{algorithm}[t]
  \caption{Logographic Encoder}\label{alg:encode}
  \begin{algorithmic}[1]
    \Procedure{Encode}{$W, C, m, \mathfrak{D}_{trgt}, \eta, f_{ct}$} 
      \State $X \gets $ EMBEDDING$(W)$
      \State $\mathfrak{D} \gets 0$
      \State $k \gets \lceil \sqrt[m]{|X|} \rceil $ \label{ln:k}
      \While{$\mathfrak{D} <\mathfrak{D}_{trgt}$}
        \State $Q \gets $ QUANTIZE$(X, m, k)$
        \State $ \mathfrak{D} \gets e^{1-\frac{|W|}{|Q|}}$ \label{ln:D}
        \State $k \gets k + \eta $ 
      \EndWhile
      \For {$w$ \textbf{in} $C$}
        \If {$w.f < f_{ct}$}
          \State DECOMPOSE$(w, Q)$
        \EndIf
      \EndFor
    \EndProcedure
  \end{algorithmic}

\end{algorithm}

\begin{table}[t]
\centering
\setlength\tabcolsep{4pt}
\caption{\label{tab:alg_params} Terms used in the proposed system }
\begin{tabular}{|l|l|}
\hline \bf Term &\bf Meaning \\
\hline
$W$ & the set of all words \\
$X$ & the set of all word vectors \\
$Q$ & the set of all code vectors \\
$m$ & the number of partitions \\
$k$ & the number of clusters per subspace\\
$\mathfrak{D}$ & the degree of distinctness\\
$\eta$ & the learning rate \\
$C$ & the corpus of all sentence pairs \\
$\mathfrak{D}_{trgt}$ & the distinctness objective \\
$f_{ct}$ & the cutoff frequency \\
\hline
\end{tabular}

\end{table}

\section{Evaluation}

In this section, we evaluate the translation performance of the proposed logographic subword model in both the RNN and CNN sequence-to-sequence translators.

\subsection{Experiment Setup}

The United Nation Parallel Corpus \cite{Ziemski:16} is used for model training and testing. English (alphabetic language) and Chinese (logographic language) translations are evaluated for both the RNN and CNN translators. The first 500000 pair-wise aligned sentences are taken from the corpus's training dataset, among which sentences longer than 40 words are not selected, (in order to simplify the model complexity brought unnecessarily to the experiments). This results in 259644 pair-wise aligned sentences for English and Chinese. To be consistent with the training dataset, sentences longer than 40 words are also not selected from the corpus's development and test dataset, and there are remaining 1928 and 1948 (out of 4000) pair-wise aligned sentences for development and test respectively. In total, there are 61168 and 31700 tokens for English and Chinese respectively in those sentences. 

We evaluate the baseline model, and our logographic subword model with a LOPQ and DAPQ quantizer (L-SW-LO and L-SW-DA respectively). In the encoder of the logographic subword model, GloVe \cite{Pennington:14} is used to construct embeddings of logograms, wherein every word vector has 6 dimensions. The LOPQ and DAPQ quantizers are both outputting 3-dimension code vectors. Code vectors are formatted to abstract subword (symbol) vectors as illustrated in Figure \ref{fig:examples}. In these experiments, $\mathfrak{D}_{trgt}$ are all set to $1$, thus there exists a one-to-one correspondence between a logographic word and the vector of abstract subwords (code symbols), and the encoding is fully reversible.  

The RNN seq2seq model is trained and evaluated on Nematus \cite{Sennrich:17} and Theano \cite{Al-Rfou:16}. It is a 2-layer bidirectional seq2seq model with each GRU cell having 1024 units. Input embedding to both encoder and decoder cells have 512 dimensions. In the experiments with L-SW-LO and L-SW-DA, every infrequent word is replaced by 3 code symbols. Since infrequent words rarely appear in sentences, the sentence lengths do not increase substantially. Thus, in those experiments, the cut-off lengths are set to 50. Theano is used to evaluate the proposed approach with the RNN seq2seq model. A Nvidia GTX 1080ti GPU is assigned to each translation task for both training and inference. 

The CNN seq2seq model is trained and evaluated on Fairseq \cite{Gehring:17,Gehring:16} and Torch \cite{Collobert:11}. Both the encoder and decoder in the CNN seq2seq model are fully convolutional. The target dictionary size decides the model complexity and how the hyperparameters are set. Since there are larger target dictionaries in the experiments with the baseline model, we define 6 convolutional layers with 768 channels in the first four layers and 1024 channels in the last two. The kernel sizes are set to 3 for the first five layers and 1 for the last layer. Since there are smaller dictionaries for L-SW-LO and L-SW-DA, we define 4 convolutional layers with 384 channels in the first two layers and 512 channels in the last two. The kernel sizes are set to 5 for the first two layers, 3 for the third layer and 1 for the last layer. Input embedding to both encoder and decoder have 128 dimensions. Cut-off lengths are set to 50 for all language pairs. Torch is used to evaluate the proposed approach with the CNN seq2seq model. Two Nvidia GTX 1080ti GPUs are assigned to each translation task for both training and inference.

\subsection{Results}

We examine how our approach improves the performance of both the RNN seq2seq model \cite{Cho:14,Bahdanau:14} and the CNN seq2seq model \cite{Gehring:17}. During encoding, every logographic word in those sentences is replaced by their corresponding abstract subword (symbol) vectors. New dictionaries are created from those encoded sentences. This results in much smaller dictionaries for Chinese (as indicated by the TrgtV column in Table \ref{tab:metrics}). We then train both the RNN and CNN sequence-to-sequence models with them. 

During testing, BLEU scores are evaluated by comparing the predicted sentences with the candidates. It is observed that BLEU scores are comparable to the baseline model and slightly better in the Chinese-English translation. 

Model sizes (NN), training time (T) and inference time (t) are measured as shown in Table \ref{tab:metrics}. We observe that in the English-to-Chinese translation task, the model sizes are reduced by $37\%$ and $77\%$ respectively with RNN and CNN; the training time is $11\%$ and $73\%$ shorter; and the inference time is nearly halved in RNN and $36\%$ shorter in CNN.

\begin{table}[t]
\small

\setlength\tabcolsep{4pt}

\caption{\label{tab:metrics}  ``PrePr'': preprocessing method;``TrgtV'': the number of tokens in the target dictionary; ``NN'': the sizes of the neural networks; Per RNN, ``T'' is the elapsed time of 15 epochs of training on one GTX1080TI GPU; per CNN, ``T'' is the elapsed time of 18 epochs of training on two GTX1080TI GPUs. ``t'' is the average inference time for the translation per each sentence. ``L-SW-LO'' stands for the logographic subword model with LOPQ encoding. ``L-SW-DA'' stands for the logographic subword model with DAPQ encoding. }
\subfloat[RNN Seq2Seq Model]{
\begin{tabular}{|l|l|rrrrr|}
\hline \bf L &  \bf PrePr & \bf TrgtV & \bf NN(Mb) & \bf T(hr) & \bf t(ms) & \bf BLEU \\ \hline

 \multirow{3}{*}{\rotatebox{90}{EN-ZH}} &- &  31702 &  466.6 &  16.45 &241.79& \bf 50.31 \\

&L-SW-LO &  4997 &  236.4 &  15.39 & 128.53& 49.95 \\
&L-SW-DA &  \bf 3492 &  \bf 228.1 &  \bf 13.45 & \bf 124.43& 49.63 \\
 \cline{1-7}
 \multirow{3}{*}{\rotatebox{90}{ZH-EN}} &- & 61170 & 527.0 & 16.81 & 320.84&48.73 \\

&L-SW-LO & 7844 & 253.7 & \bf 14.93 &137.58& 48.59 \\
&L-SW-DA & \bf 6632 & \bf 234.3 & 14.94 &\bf 130.92 & \bf 49.23 \\
 
\hline
\end{tabular}
}
\quad
\subfloat[CNN Seq2Seq Model]{
\begin{tabular}{|l|l|rrrrr|}
\hline \bf L &  \bf PrePr & \bf TrgtV & \bf NN(Mb) & \bf T(hr) & \bf t(ms) & \bf BLEU \\ \hline
 \multirow{3}{*}{\rotatebox{90}{EN-ZH}} &- & 31703 &  501.8 &  2.54 &35.93 & \bf 49.22 \\
&L-SW-LO & 4998 & 112.2 & 0.98 & \bf 21.05 & 47.07 \\
&L-SW-DA & \bf 3493 & \bf 107.8 & \bf 0.87 & \bf 21.05 & 47.47\\
 \cline{1-7}
 \multirow{3}{*}{\rotatebox{90}{ZH-EN}} &- & 61171 & 532.8 & 3.06 & 34.91&48.21 \\
&L-SW-LO & 7845 & 115.5 & 1.01 & \bf 21.05 & 47.39 \\
&L-SW-DA & \bf 6633 & \bf 111.4 & \bf 1.00 & 22.07& \bf 48.22 \\

\hline
\end{tabular}
}

\end{table}

We compare the logographic subword approaches with the subword BPE approach for Chinese-English translation (Table \ref{tab:metrics-sw}). BPE cannot identify subwords in Chinese logograms. Consequently, it hurts the final BLEU score. Our approaches fit better to logographic languages in this direction. In the direction from English to Chinese, we did not observe the proposed models are better than BPE-based models. 

\begin{table}[t]
\small
\centering
\caption{\label{tab:metrics-sw} In the task of Chinese-English translation, the BPE approach cannot identify subwords in Chinese logograms and consequently hurts the final BLEU score.
}
\setlength\tabcolsep{4pt}

\begin{tabular}{|l|rrrr|}
\hline \bf PrePr &  \bf NN(Mb) & \bf T(hr) & \bf t(ms) & \bf BLEU \\ \hline
SW-BPE   & 347.2 & \bf 14.59 &169.92& 47.66 \\
L-SW-LO  & 253.7 & 14.93 &137.58& 48.59 \\
L-SW-DA & \bf 234.3 & 14.94 &\bf 130.92 & \bf 49.23 \\
\hline
\end{tabular}

\end{table}

\begin{figure}[t]
\centering
\subfloat{
\includegraphics[width=0.25\linewidth]{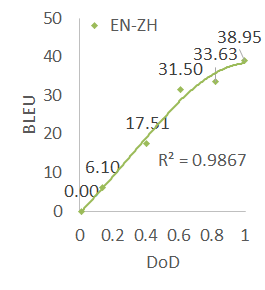}
}
\quad
\subfloat{
\includegraphics[width=0.25\linewidth]{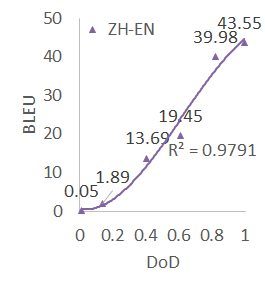}
}

\caption[Experiment - $\mathfrak{D}$ and $BLEU$]{There exists a correlation between the \textit{degree of distinctness} $\mathfrak{D}$ and the BLEU score at $\mathfrak{D} =  e^{0.5(1-\frac{|W|}{|Q|})}$ on a CNN seq2seq model. The BLEU score is polynomially proportional to $\mathfrak{D}$ for both language pairs. }
\label{fig:CNN-DoD-BLEU}
\end{figure}

Reversibility is critical to translation performance. We explore various values of DoD $\mathfrak{D}$ (at cut-off frequency $f_{ct} = \infty$) and evaluate the translation performance and model sizes in the CNN seq2seq model (Figure \ref{fig:CNN-DoD-BLEU}). Experiments are conducted for language pairs with the same hyper-parameters except different $\mathfrak{D}$ values. As expected, the BLEU score increases as $\mathfrak{D}$ increases and reaches the maximum when $\mathfrak{D}$ equals to $1$, since the encoding is fully reversible at this point. Figure \ref{fig:CNN-DoD-BLEU} demonstrates the correlation between DoD and the BLEU score at $\mathfrak{D} =  e^{0.5(1-\frac{|W|}{|Q|})}$ ($b = 0.5$ in Equation \ref{eq:D}); BLEU is proportional to $\mathfrak{D}$ in both translation tasks. 

\section{Conclusion}
A logographic subword model is proposed, with an encoder and decoder transforming logograms to abstract subwords and subwords to logograms. The encoder quantizes and decomposes the embeddings of logograms. By sharing common abstract subwords (code symbols), quantization reduces the dictionary size without sacrificing translation performance. A new metric, \textit{degree of distinctness}, is proposed to quantify the effect of distinctness and reversibility. %Infrequent words are taken into account to avoid unnecessary elongation of the source and target sentences. 

The proposed approach has been shown with experiments to reduce model sizes as well as shorten training and inference time for both RNN and CNN sequence-to-sequence models. It is promising for reducing the complexity of other computationally expensive NLP problems with potential impact on large-dictionary real-time offline applications such as translation or dialog systems on offline mobile platforms. 

As future work, we will build a faster and more accurate encoder and decoder, and explore the use of abstract subwords in other logographic languages such as Chorti and Demotic. Those are important since the proposed techniques help to understand different logographic languages from the subword perspective.

\bibliographystyle{unsrt}
\bibliography{references}

\end{document}